\pdfoutput=1 
\documentclass{article}
\usepackage{arxiv}

\usepackage{url}
\usepackage{graphicx}
\usepackage{multirow}
\usepackage[utf8]{inputenc}
\usepackage{multicol}
\usepackage{booktabs}
\usepackage{amsmath}
\usepackage{natbib}
\usepackage{url}

\newcommand{\ent}[1]{\textit{#1}}

\begin{document}

\title{HunFlair: An Easy-to-Use Tool for State-of-the-Art Biomedical Named Entity Recognition}
\author{Leon Weber\,$^{\text{1, 2},*,\dagger}$,\\
Mario Sänger\,$^{\text{1},*,\dagger}$,\\
Jannes Münchmeyer\,$^{\text{1,3}}$,\\
Maryam Habibi\,$^{\text{1}}$,\\
Ulf Leser\,$^{\text{1}}$ and\\
Alan Akbik\,$^{\text{1}}$ \\
\small
$^{\text{\sf 1}}$Computer Science Department, Humboldt-Universität zu Berlin\\
\small
 $^{\text{\sf 2}}$Mathematical Modelling of Cellular Processes, Max Delbrück Center\\\small for Molecular Medicine in the Helmholtz Association\\
\small
 $^{\text{\sf 3}}$Seismology, GFZ German Research Centre for Geosciences\\
\small 
$^\ast$To whom correspondence should be addressed.\\
\small
$^\dagger$Both authors contributed equally.}

\begin{quote}
\abstract{
\textbf{Summary:} Named Entity Recognition (NER) is an important step in biomedical information extraction pipelines. Tools for NER should be easy to use, cover multiple entity types, highly accurate, and robust towards variations in text genre and style. To this end, we propose \ent{HunFlair}, an NER tagger covering multiple entity types integrated into the widely used NLP framework \ent{Flair}. \ent{HunFlair} outperforms other state-of-the-art standalone NER tools with an average gain of 7.26 pp over the next best tool, can be installed with a single command and is applied with only four lines of code.  \\
\textbf{Availability:} \ent{HunFlair} is freely available through the \ent{Flair} framework under an MIT license: \url{https://github.com/flairNLP/flair/blob/master/resources/docs/HUNFLAIR.md} and is compatible with all major operating systems.\\
\textbf{Contact:} \url{{weberple,saengema,alan.akbik}@informatik.hu-berlin.de}\\
online.}
\end{quote}

\maketitle

\section{Introduction}
Recognizing biomedical entities such as genes, chemicals or diseases in unstructured scientific text is a crucial step for  biomedical information extraction pipelines.
Tools for named entity recognition (NER) are typically trained and evaluated on rather small gold standard data sets. 
However, they are applied "in the wild", i.e., to a much larger collection of texts, often varying in topic, entity distribution, genre (e.g. patents vs. scientific articles) and text type (e.g. abstract vs. full text).
This mismatch between evaluation and application scenario can lead to severe drops in performance.
To address this, we recently released the \ent{HUNER} tagger~\citep{weber-munchmeyer-rocktaschel-habibi-leser-2019-huner} that was trained on a large collection of biomedical NER datasets, leading to a much better performance on unseen corpora compared to models trained on a single corpus.
However, \ent{HUNER} relies on a Docker installation and uses a client-server architecture that cannot be easily connected to any of the major NLP frameworks for further processing of the input text.
Moreover, it doesn't incorporate pretrained language models although these were the basis for many recent breakthroughs in NLP research~\citep{akbik-etal-2019-flair}.

In this work, we train, evaluate and make available NER models for the five entity types \ent{Cell Line}, \ent{Chemical}, \ent{Disease}, \ent{Gene} and \ent{Species} within the easy-to-use \ent{Flair} framework~\citep{akbik-etal-2019-flair}. 
Technically, the models combine the insights from \citet{weber-munchmeyer-rocktaschel-habibi-leser-2019-huner} and \citet{akbik-etal-2019-flair} by merging character-level pretraining and joint training on multiple gold standard corpora, which leads to strong gains over other state-of-the-art standalone NER tools. 
Integration into \ent{Flair} adds very simple usage even for non-experts: \ent{HunFlair} can be installed with a single command and applied with only a few lines of code.
Additionally, we integrate 23 biomedical NER corpora into \ent{HunFlair} using a consistent format, which enables researchers and practitioners to rapidly train their own models and experiment with new approaches.

\section{HunFlair}
HunFlair is based on and integrated into the \ent{Flair} NLP framework. 
\ent{Flair} is designed to allow intuitive training and distribution of sequence labeling, text classification and language models, achieving state-of-the-art performance in several NLP research challenges. 
It allows researchers to “mix and match” various types of character, word and document embeddings with little effort.

\begin{figure}[t]
  \centering
  \includegraphics[width=\linewidth]{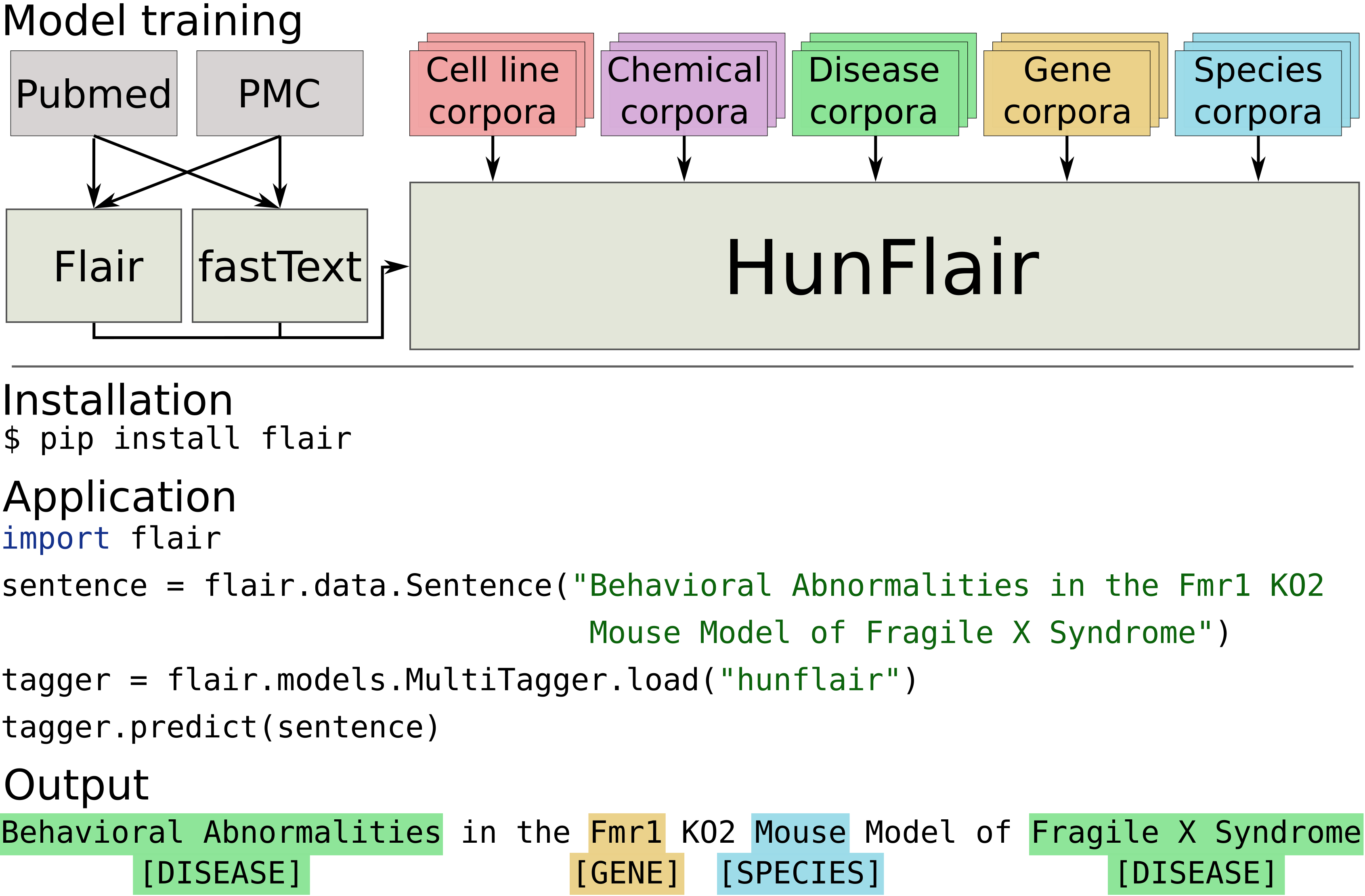}
  \caption{Overview of the \ent{HunFlair} model trained on 23 biomedical NER data sets in total, its installation and a sample application with annotated output.}
  \label{fig:hunflair} 
\end{figure}

Figure~\ref{fig:hunflair} illustrates the architecture of \ent{HunFlair}.
At the core, it relies on a Flair character-level language model trained on roughly 24 million abstracts of biomedical articles from PubMed and 
3 million full texts originating from PMC as well as fastText word embeddings~\citep{bojanowski-etal-2017-enriching}.  
Prediction of named entities is performed by a BiLSTM-CRF model.
Analogously to HUNER, we train distinct models for each entity type using the union of all training sets of all gold standard NER corpora with this type to improve performance over text genres and biomedical sub-domains. 
See SM~1 for details of the training process.

\section{Results}
We compare \ent{HunFlair} to two types of competitors: Other off-the-shelf biomedical NER tools with the ambition to reduce effort for being used, and other recent research prototypes, which often achieve better performance but require considerable more work.

\subsection{Comparison to off-the-shelf NER tools}
We compare the results of \ent{HunFlair} to those of other state-of-the-art standalone biomedical NER tools on the complete CRAFT~\citep{bada2012concept}, BioNLP13 Cancer Genetics~\citep{pyysalo-etal-2013-overview} and PDR~\citep{kim-wonjun-lee-2019-corpus} corpora. None of these was used in the training of neither \ent{HunFlair} nor the competitor tools.
We compare against \ent{SciSpacy}~\citep{neumann-etal-2019-scispacy}, \ent{HUNER}~\citep{weber-munchmeyer-rocktaschel-habibi-leser-2019-huner}, \ent{tmChem}~\citep{leaman2015tmchem}, \ent{GNormPlus}~\citep{wei2015gnormplus} and \ent{DNorm}~\citep{leaman-2013-dnorm}.
As \ent{SciSpacy} comes with several models for each entity type, we report the best performance among all of those models that were not trained on the evaluation corpus.
Results can be found in Table~\ref{tab:cross_corpus_results}.
\ent{HunFlair}  outperforms all competitors in all but one comparisons, with an average gain of 7.26 pp in F1.
Note, that we evaluate using the gold spans annotated in the original corpora, allowing for a fair comparison across different pre-processing procedures.
See SM~2 for the details on the evaluation protocol and a discussion of the results.

\subsection{Comparison to research prototypes}
We compare \ent{HunFlair} to the reported scores of the state-of-the-art models \ent{BioBERT}~\citep{lee-et-al-2019-biobert}, \ent{SciBERT}~\citep{beltagy-etal-2019-scibert}, \ent{CollaboNet}~\citep{Yoon2019} and \ent{SciSpacy}~\citep{neumann-etal-2019-scispacy}.
The results can be found in SM~3.
\ent{HunFlair} sets the new state-of-the-art on one corpus and performs on-par on the others.
We also investigate the effect of pretraining on multiple gold standard corpora, by comparing \ent{HunFlair} to a non-pretrained version on all 23 NER corpora.
On average, finetuning improves results on all entity types with the improvements in F1 ranging from $0.8$ pp for chemicals to $4.75$ pp for cell lines.
The full results per corpus are provided in SM~4.

\section{Conclusion}
We proposed \ent{HunFlair}, a state-of-the-art biomedical NER tagger.
\ent{HunFlair}, which builds on pretrained domain-specific language models, outperforms other tools on unseen corpora, often by a large margin. 
It is easy to install and use and comes along with 23 biomedical NER corpora in a single format to ease future research.

\begin{table}[h]
\scriptsize
\caption[]{F1-scores of several off-the-shelf biomedical NER tools on three unseen corpora. We distinguish entity types Chemical (Ch), Disease (D), Gene (G) and Species (S). The best results are in bold. Misc displays the results of multiple taggers: tmChem for Chemical, GNormPus for Gene and Species, and DNorm for Disease.}
\label{tab:cross_corpus_results}
\centering
\begin{tabular}{lrrrrrrrrrr}
\toprule
\multicolumn{1}{c}{} & \multicolumn{3}{c}{CRAFT} && \multicolumn{4}{c}{BioNLP CG} && \multicolumn{1}{c}{PDR}\\
& Ch & G & S && Ch & D & G & S && D \\
\cmidrule{2-4} 
\cmidrule{6-9}
\cmidrule{11-11}
Misc & 
    42.88 & 64.93 & 81.15 && 
    72.15 & 55.64 & 68.97 & \textbf{80.53} && 
    80.63
    \\

SciSpacy & 
    35.73 & 47.76 & 54.21 && 
    58.43 & 56.48 & 66.18 & 57.11 &&
    75.90
    \\
HUNER & 42.99 & 50.77 & 84.45 &&
        67.37 & 55.32 & 71.22 & 67.84 &&  
        73.64
       \\
HunFlair & \textbf{59.69} & \textbf{72.19} & \textbf{85.05} &&
         \textbf{81.82} & \textbf{65.07} & \textbf{87.71} & 76.47 &&
         \textbf{83.44}
\\

\bottomrule
\end{tabular}
\end{table}

\section*{Funding}
Leon Weber and Jannes Münchmeyer are funded by the Helmholtz Einstein International Berlin Research School in Data Science (HEIBRiDS). Maryam Habibi is funded by the German Research Council, grant LE-1428/7-1.

\bibliographystyle{natbib}
\bibliography{references}

\end{document}


\maketitle

\section{Training of HunFlair}
The training of \ent{HunFlair} is a two-step process.
First, the required word embeddings are trained on a large unlabeled corpus, which are then used in the training of the NER tagger on multiple manually labeled NER corpora. 

\subsection{Embeddings}
\label{sec:embeddings}
We use two types of word embeddings for \ent{HunFlair}, (I) \ent{Flair} embeddings based on a character-level language model (LM) and (II) \ent{fastText} embeddings. 
We trained the Flair LM on a corpus of roughly 3 million full texts from the PubmedCentral BioC text mining collection\footnote{\url{ftp://ftp.ncbi.nlm.nih.gov/pub/wilbur/BioC-PMC/}, Version of 2019/05/24} and 25 million abstracts of PubMed articles\footnote{ftp://ftp.ncbi.nlm.nih.gov/pubmed/baseline, Version of 2019/12/16A}, yielding a corpus of roughly 14 billion tokens, which we divide into 1500 splits.
For the training of fastText, we used the same corpus, which we enriched with the text of 6,062,172 wikipedia articles\footnote{\url{https://dumps.wikimedia.org/enwiki/latest/enwiki-latest-pages-articles.xml.bz2}, Version of 2020/05/06}, adding another 2.6 billion tokens.
For the Flair embeddings, we use a single-layer LSTM with a hidden size of $2048$ for each direction.
Both LSTMs are trained with a sequence length of $300$, a batch size of $256$ and a split-wise patience for the learning rate annealing of $100$.
For the fastText embeddings, we train a skip-gram model with $200$ dimensions and sample 10 negative examples per step.
The rest of the hyperparameters is left at their default value.

\subsection{Gold standard NER pre-training}
In order to have a broad data basis, we harmonize 23 manually-curated, biomedical NER corpora for the training of \ent{HunFlair}. 
The corpora include patents, abstracts and full-texts from scientific articles and are annotated with a variety of entity types. 
Table~SM~\ref{tab:hunflair_corpora} gives an overview about the included corpora and highlights important statistics.
We use the sentence splitter and a modified version of the tokenizer of the \textit{en\_core\_sci\_sm} model of \ent{scispacy}\footnote{\url{https://s3-us-west-2.amazonaws.com/ai2-s2-scispacy/releases/v0.2.5/en_core_sci_sm-0.2.5.tar.gz}}~\citep{neumann-etal-2019-scispacy}.
We train distinct models for each entity type, i.e. cell lines, chemicals, disease, gene / proteins and species, to achieve high quality results. 
For each type we only use corpora that contain annotations for the respective entity type to learn a type-specific model.
We re-use the splits introduced by \ent{HUNER} \citep{weber-munchmeyer-rocktaschel-habibi-leser-2019-huner} to form a training and validation split for each data set. 
Our training sets are built by taking the union of the \ent{HUNER} train and test splits of each data set.
The validation sets are given by the union of all \ent{HUNER} validation splits.
The former is used to train the models and the latter to select the best performing model.

We apply a bidirectional LSTM-CRF neural network to model the recognition of named entities as sequence labeling task.
We represent input words using the \ent{HunFlair} language model and fastText embeddings learned on in-domain texts (see Section~\ref{sec:embeddings}).
Building on this, a single layer Bi-LSTM with a hidden size of 256 is used to process the input sequence.
Prediction of the output sequence, i.e. one IOBES label per word, is done using a CRF in the final layer.
All models are trained for 200 epochs with an batch size of 32, an initial learning rate of $0.1$, dropout of $0.5$ and a patience of $3$. 

\begin{landscape}

  \begin{table}  
  \centering
  \footnotesize
  
  \caption{Overview of the 23 biomedical NER corpora used to train HunFlair. For each corpus we report the text genre (patent (P) / scientific articles (SA)), text type (abstract (A) / full-text (FT)) as well as number of sentence, token, entity annotation statistics.}
  \label{tab:hunflair_corpora}
 
    \begin{tabular}{lllrrlrr}
        \toprule
      Corpora & Genre & Type & \mc{1}{l}{Sentences} & \mc{1}{l}{Tokens} & \mc{1}{l}{Entity Type} & Annotations & \mc{1}{l}{Unique Ann.} \\ 
      \midrule  

      BioCreative II GM \tcite{Smith2008-zi} & SA & A & 20,744 & 545,966 & Genes / Proteins & 24,453 & 16,046 \\ 
    
      BioCreative V GPRO \tcite{perez2017evaluation} & P & A & 35,277 & 1,558,687 & Genes / Proteins & 13,125 & 5,662 \\  

      BioInfer \tcite{Pyysalo2007-ah} & SA & A & 1,138 & 37,135 & Genes / Proteins & 4,408 & 1,357 \\  

      CellFinder \tcite{Neves2012-cq} & SA & FT & 2,211 & 70,286 & Cell Lines & 367 & 63 \\  

       &  &  &  &  & Genes / Proteins & 1,572 & 706 \\  

       &  &  &  &  & Species & 462 & 43 \\  

      CDR \tcite{Li2016-yh} & SA & A & 14,464 & 345,648 & Chemicals & 15,828 & 2,712 \\  

       &  &  &  &  & Diseases & 12,931 & 3,281 \\  

      CHEMDNER patent \tcite{Krallinger2015-od,Krallinger2015-wu} & P & A & 48,744 & 1,558,182 & Chemicals & 65,238 & 20,529 \\  

      CHEBI \tcite{Shardlow2018-dw} & P & FT & 13,088 & 423,731 & Chemicals & 24,124 & 6,816 \\  

       &  &  &  &  & Genes / Proteins & 7,140 & 1,871 \\  

       &  &  &  &  & Species & 3,841 & 884 \\  

      CHEMDNER \tcite{Krallinger2015-wu} & SA & A & 87,550 & 2,431,366 & Chemicals & 83,058 & 20,470 \\  

      CLL \tcite{Kaewphan2016-yw} & SA & A, FT & 201 & 7,689 & Cell Lines & 341 & 309 \\  

      DECA \tcite{Wang2010-kb} & SA & A & 5,454 & 147,874 & Genes / Proteins & 6,261 & 2,187 \\  

      FSU-PRGE \tcite{Hahn2010-bd} & SA & A & 36,216 & 985,598 & Genes / Proteins & 59,521 & 15,912 \\  

      Gellus \tcite{Kaewphan2016-yw} & SA & A, FT & 11,809 & 312,699 & Cell Lines & 650 & 210 \\  

      IEPA \tcite{Ding2002-hl} & SA & A & 486 & 16,590 & Genes / Proteins & 1,117 & 139 \\  

      JNLPBA \tcite{Kim2004-fb} & SA & A & 18,535 & 532,777 & Cell Lines & 3,831 & 2,250 \\  

       &  &  &  &  & Genes / Proteins & 30,263 & 8,964 \\  

      Linneaus \tcite{Gerner2010-uw} & SA & FT & 17,593 & 504,261 & Species & 2,724 & 339 \\  

      LocText \tcite{Goldberg2015-md} & SA & A & 945 & 24,178 & Genes / Proteins & 1,930 & 717 \\  

       &  &  &  &  & Species & 276 & 37 \\  

      miRNA \tcite{Bagewadi2014-op} & SA & A & 2,456 & 64,897 & Diseases & 2,032 & 586 \\  

       &  &  &  &  & Genes / Proteins & 944 & 345 \\  

       &  &  &  &  & Species & 676 & 45 \\  

      NCBI Disease \tcite{Dogan2014-if} & SA & A & 7,308 & 179,849 & Diseases & 6,861 & 2,137 \\  

      OSIRIS \tcite{Furlong2008-ia} & SA & A & 1,072 & 31,020 & Genes / Proteins & 957 & 355 \\  

      S800 \tcite{Pafilis2013-vv} & SA & A & 6,421 & 165,451 & Species & 3,734 & 1,576 \\  

      SCAI Chemical \tcite{Kolarik2008-ln} & SA & A & 940 & 30,808 & Chemicals & 1,314 & 797 \\  

      SCAI Disease \tcite{Gurulingappa2010-xa} & SA & A & 4,351 & 113,541 & Diseases & 2,241 & 1,003 \\  

      Variome \tcite{Verspoor2013-vy} & SA & FT & 6,155 & 180,237 & Diseases & 5,925 & 475 \\  

       &  &  &  &  & Genes / Proteins & 4,552 & 529 \\  

       &  &  &  &  & Species & 182 & 8 \\ 
      \bottomrule  

    \end{tabular}
  \end{table}
\end{landscape}

\section{Evaluation against off-the-shelf tools}
The evaluation of \ent{HunFlair} and its competitor biomedical NER tools is performed using 
the three corpora, CRAFT~\citep{bada2012concept}, BioNLP13 Cancer Genetics~\citep{pyysalo-etal-2013-overview} and plant-disease-relations (PDR)~\citep{kim-wonjun-lee-2019-corpus}. 
For the comparison with \ent{SciSpacy}~\citep{neumann-etal-2019-scispacy}, we use the models \ent{en\_ner\_craft\_md}, \ent{en\_ner\_jnlpba\_md}, \ent{en\_ner\_bc5cdr\_md}, and \ent{en\_ner\_bionlp13cg\_md}\footnote{Note, that we don't compare against the more general \ent{SciSpacy} models (e.g. \ent{en\_core\_sci\_md} or \ent{en\_core\_sci\_lg}), since they do not provide entity types out-of-the-box.}. 
However, when evaluating on a corpus which was used to train the specific \ent{SciSpacy} model, we excluded the respective model and report the best score of the remaining models to retain a fair comparison.
Due to this, neither \ent{HunFlair} nor any of the competitor tools are trained on any of the corpora, hence the evaluation setting is similar to an application to completely unseen text. 
Table~SM~\ref{tab:eval_corpora} highlights statistics of the used corpora.

\begin{table*} 
  \centering
  \footnotesize
  \setlength{\tabcolsep}{5pt}
  \caption{Overview of the gold standard NER corpora used to evaluate \ent{HunFlair} and the competitor models in an cross-corpus setting. For each corpus we report the number of sentences and tokens as well as entity annotation statistics.}
  \label{tab:eval_corpora}
  
    \begin{tabular}{lrrlrr}
    \toprule
      Corpora & \mc{1}{l}{Sentences} & \mc{1}{l}{Tokens}& Entity Type & \mc{1}{l}{Annotations} & \mc{1}{l}{Unique} \\ 
      \midrule  

      BioNLP2013-CG & 5,994 & 157,109 & Chemicals & 2,405 & 841 \\
        \tcite{pyysalo-etal-2013-overview}  
        & & & Diseases & 2,604 & 624 \\
        & & & Genes / Proteins & 7,908 & 2,057 \\
        & & & Species & 1,801 & 306 \\

      CRAFT & 26,589 & 776,028 & Chemicals & 6,780 & 1,031 \\  
        \tcite{bada2012concept}
        & & & Genes / Proteins & 23,578 & 2,330 \\  
        & & & Species & 10,465 & 354 \\

      Plant-Disease (PDR) & 1,780 & 49,392 & Diseases & 1,298 & 477 \\  

      \tcite{kim-wonjun-lee-2019-corpus}  &  &  & \\ 
      \bottomrule  

    \end{tabular}
\end{table*}

We report F1 scores for all considered methods and tools. 
%
We designed our evaluation to minimize the assumptions made about the preprocessing of the input texts, especially with respect to tokenization and sentence splitting. 
Each model is given the complete abstract resp. full-text of the scientific article or patent as input for which it executes its own pre-processing pipeline.
The predictions of each model are represented by text offsets.
To calculate the evaluation scores, we use the gold standard text offsets and match them with the predicted offsets. 
We consider any predicted span as true positive that either matches exactly one gold standard annotation or differs only by one character either at the end or at the beginning.
This accounts for the fact that the methods have differences in their processing of special characters, leading to small deviations in token off-sets.

Note, that this evaluation protocol differs substantially from the one used in~\citet{weber-munchmeyer-rocktaschel-habibi-leser-2019-huner}, where homogeneously preprocessed versions of the corpora were used for evaluation, leading to different offsets in many cases.
Additionally, \ent{HUNER} only outputs the entities as extracted from the tokenized text, losing non-ascii symbols and whitespace in the process.
Thus, to align the predicted entities to the input text in the present evaluation, we try to match the predicted entity strings to the original text by using fuzzy matching.
These are two important reasons for the results for Gene on the Craft corpus are much worse than those reported in~\citet{weber-munchmeyer-rocktaschel-habibi-leser-2019-huner}.
This is supported by the fact that the difference between results from~\citet{weber-munchmeyer-rocktaschel-habibi-leser-2019-huner} and those reported diminishes, when counting any overlap between predicted and annotated spans as a true positive (see Table~SM~\ref{tab:cross_corpus_results}). 
We noticed that for some combinations of model and corpus \ent{SciSpacy} predicts wrong entity boundaries in a large number of cases, leading to strikingly different results in the any-overlap evaluation and the more strict one.
Nevertheless, also under this evaluation protocol, \ent{HunFlair} performs better than all competitors except for Species on the BioNLP corpus.

\begin{table}[ht!]
\scriptsize
\caption[]{ Cross corpus evaluation of off-the-shelf BioNER tools for the entity types Chemical (Ch), Disease (D), Gene (G) and Species (S) counting any overlap between the predicted and annotated span as a true positive. All scores are F1-measures and the best results are in bold. Delta shows the improvement over the more strict evaluation reported in the main text (Table 1). Misc displays the results of multiple taggers: tmChem for Chemical, GNormPus for Gene and Species, and DNorm for Disease.}
\label{tab:cross_corpus_results}
\centering
\begin{tabular}{lrrrrrrrrrr}
\toprule
\multicolumn{1}{c}{} & \multicolumn{3}{c}{CRAFT} && \multicolumn{4}{c}{BioNLP CG} && \multicolumn{1}{c}{PDR}\\
& Ch & G & S && Ch & D & G & S && D \\
\cmidrule{2-4} 
\cmidrule{6-9}
\cmidrule{11-11}
Misc & 
    44.86 & 67.52 & 82.02 &&
    74.36 & 60.04 & 71.06 & \textbf{84.18} &&
    86.95
    \\
$\Delta$ &
    1.98 & 2.59 & 0.87 && 
    2.21 & 4.40 & 2.09 & 3.65 &&
    6.32
    \\
    \\

SciSpacy & 
    39.78 & 54.71 & 72.02 &&
    60.65 & 61.69 & 78.77 & 65.04 &&
    83.49 
    \\
$\Delta$ &
    4.05 & 6.95 & 17.81 &&
    2.22 & 5.21 & 12.59 & 7.93 &&
    7.59
    \\
    \\
HUNER & 46.84 & 65.27 & 84.66 &&
        72.00 & 59.74 & 79.58 & 71.43 &&
        78.49
       \\
$\Delta$ &
    3.85 & 14.50 & 0.21 &&
    4.63 & 4.42 & 8.36 & 3.59 &&
    4.85
   \\ 
    \\
HunFlair & \textbf{61.99} & \textbf{80.5} & \textbf{85.46} &&
         \textbf{83.52} & \textbf{69.29} & \textbf{92.73} & 80.15 &&
         \textbf{88.64}
         \\
$\Delta$ &
    2.16 & 6.99 & 0.42 &&
    1.70 & 4.22 & 5.02 & 3.74 &&
    5.20
    \\
         
\bottomrule
\end{tabular}
\end{table}

\section{Evaluation against state-of-the-art models}
\begin{table*}[htp]
\caption{Comparison with the reported results of state-of-the-art models for BioNER. Scores are macro-averaged F1 and best results are printed in bold. 'HunFlair (no)' refers to the HunFlair model without pretraining on goldstandard corpora.}
\label{tab:sota}
\centering
\begin{tabular}{lrrr}
\toprule
            & JNLPBA (Gene) & BC5CDR & NCBI \\
            \midrule
SciBERT     & 77.28 & 90.01 & 88.57\\
BioBERT v1.1& 77.49 & 89.76 & \textbf{89.71}\\
CollaboNET  & \textbf{78.58} & 87.68 & 88.60\\
SciSpacy    & - & 83.92 & 81.56\\
\midrule
HunFlair    & 77.6 & 89.65 & 88.65\\
HunFlair (no)    & 77.78 &\textbf{90.57} & 87.47\\
\bottomrule

\end{tabular}
\end{table*}
We compare \ent{HunFlair} to the reported scores of the state-of-the-art models \ent{BioBERT}~\citep{lee-et-al-2019-biobert}, \ent{SciBERT}~\citep{beltagy-etal-2019-scibert}, \ent{CollaboNet}~\citep{Yoon2019} and \ent{SciSpacy}~\citep{neumann-etal-2019-scispacy} on JNLPBA (only using Gene annotations), NCBI Disease and CDR.
The results can be found in Table~SM~\ref{tab:sota}.
\ent{HunFlair} sets the new state-of-the-art on one corpus and performs on-par on the others.
For this experiment, we used the large BioWordVec embeddings~\footnote{\url{https://ftp.ncbi.nlm.nih.gov/pub/lu/Suppl/BioSentVec/BioWordVec_PubMed_MIMICIII_d200.vec.bin}}~\citep{Chen2018} and remove the three evaluation corpora from the pretraining set.

\section{Effects of pretraining}
We investigate the effects of pretraining our tagger on multiple goldstandard corpora, by comparing the pretrained tagger to a randomly initialized LSTM.
Note, that the randomly initialized LSTM still uses pretrained \ent{Flair} and \ent{fastText} embeddings.
For this experiment, we used the large BioWordVec embeddings and do not use the test portions of the corpora for pretraining.
The results can be found in Table~SM~\ref{tab:results_finetuning}.

Pretraining improves the average results for all entity types with gains ranging from $0.8$ pp for chemicals to $4.75$ pp for cell lines.
Performance improvements are mainly attributed to better recall.
In 28 of the 34 cases the recall of the pretrained model is higher than the vanilla one. 
For eight cases recall improves by over $4.0$ pp. 
This indicates that the increased amount of training data indeed leads to a better coverage of existing entities and their various surface forms as well as a higher adaptability to other biomedical subdomains.
However, there are also six cases where the $F1$ score decreases slightly (max. $1.05$ pp). 
In five out of these six cases there is a decline in precision.
Additionally, also in ten cases in which $F1$ increases, precision is lower. %
This suggests that the larger number of entities seen in training may occasionally lead to few imprecise predictions.

\begin{table*}[htp]
\footnotesize
\caption{Comparison of the tagger that was pretrained on multiple gold standard corpora (Pretrained) vs a tagger without pretraining (Vanilla). The $\Delta$-columns report the gains achieved through pretraining. }
\label{tab:results_finetuning}
\setlength{\tabcolsep}{2pt}
\begin{tabular}{lrrrrrrrrrrr}
\toprule
 & \multicolumn{3}{c}{Vanilla} && \multicolumn{3}{c}{Pretrained} \\
                             &   Prec. &   Rec. &     F1  && Prec. & Rec. & F1
                             && $\Delta$Prec. & $\Delta$Rec. & $\Delta$F1 \\
\cmidrule{2-4}                            
\cmidrule{6-8}                            
\cmidrule{10-12}                            
Cell Line & & & && & & \\
\cmidrule{1-1}
CellFinder & 0.9174 & 0.7634 & 0.8333 && 0.8983 & 0.8092 & 0.8514 && -0.0191 & 0.0458 & 0.0181\\
CLL & 0.7093 & 0.7922 & 0.7485 && 0.8158 & 0.8052 & 0.8105 && 0.1065 & 0.0130 & 0.0620\\
Gellus & 0.7818 & 0.6964 & 0.7366 && 0.9375 & 0.7895 & 0.8571 && 0.1557 & 0.0931 & 0.1205\\
 JLNPBA & 0.7456 & 0.6876 & 0.7154 && 0.7485 & 0.6661 & 0.7049 && 0.0029 & -0.0215 & -0.0105\\
 \midrule
  avg. & 0.7885 & 0.7349 & 0.7585 && 0.8500 & 0.7675 & 0.8060 && 0.0711 & 0.0433 & 0.0528\\
 & & & && & & \\
Chemical & & & && & & \\
\cmidrule{1-1}
 CDR & 0.9365 & 0.9391 & 0.9378 && 0.9394 & 0.9411 & 0.9403 && 0.0029 & 0.0020 & 0.0025\\
 CHEMDNER patent & 0.8491 & 0.9135 & 0.8801 && 0.8471 & 0.9187 & 0.8815 && -0.0020 & 0.0052 & 0.0014\\
CHEBI & 0.8006 & 0.7878 & 0.7941 && 0.8220 & 0.7786 & 0.7997 && 0.0214 & -0.0092 & 0.0056\\
CHEMDNER & 0.9319 & 0.9171 & 0.9245 && 0.9310 & 0.9198 & 0.9254 && -0.0009 & 0.0027 & 0.0009\\
SCAI Chemical & 0.8131 & 0.7307 & 0.7697 && 0.8505 & 0.8347 & 0.8425 && 0.0374 & 0.1040 & 0.0728\\
\midrule
 avg. & 0.8662 & 0.8576 & 0.8612 && 0.8780 & 0.8786 & 0.8779 && 0.0129 & 0.0246 & 0.0166\\
 & & & && & & \\
 
Disease & & & && & & \\
\cmidrule{1-1}
CDR & 0.8615 & 0.8727 & 0.8670 && 0.8488 & 0.8804 & 0.8643 && -0.0127 & 0.0077 & -0.0027\\
miRNA & 0.8318 & 0.8220 & 0.8269 && 0.8467 & 0.8769 & 0.8615 && 0.0149 & 0.0549 & 0.0346\\
NCBI Disease & 0.8583 & 0.8990 & 0.8782 && 0.8663 & 0.8815 & 0.8738 && 0.0080 & -0.0175 & -0.0044\\
SCAI Disease & 0.8159 & 0.7930 & 0.8043 && 0.8311 & 0.7972 & 0.8138 && 0.0152 & 0.0042 & 0.0095\\
Variome & 0.9147 & 0.9127 & 0.9137 && 0.9072 & 0.9163 & 0.9117 && -0.0075 & 0.0036 & -0.0020\\
\midrule
 avg. & 0.8564 & 0.8599 & 0.8580 && 0.8600 & 0.8705 & 0.8650 && 0.0117 & 0.0176 & 0.0106\\
 & & & && & & \\
 
Gene & & & && & & \\
\cmidrule{1-1}
BioCreative II GM & 0.8330 & 0.8284 & 0.8307 && 0.8372 & 0.8285 & 0.8328 && 0.0042 & 0.0001 & 0.0021\\
BioInfer & 0.8647 & 0.8351 & 0.8497 && 0.8813 & 0.8717 & 0.8765 && 0.0166 & 0.0366 & 0.0268\\
CellFinder & 0.8254 & 0.7045 & 0.7602 && 0.9050 & 0.8662 & 0.8852 && 0.0796 & 0.1617 & 0.1250\\
CHEBI & 0.7811 & 0.6667 & 0.7194 && 0.7810 & 0.7155 & 0.7468 && -0.0001 & 0.0488 & 0.0274\\
DECA & 0.7200 & 0.7388 & 0.7293 && 0.7390 & 0.7306 & 0.7348 && 0.0190 & -0.0082 & 0.0055\\
FSU-PRGE & 0.9036 & 0.9171 & 0.9103 && 0.9020 & 0.9187 & 0.9103 && -0.0016 & 0.0016 & 0.0000\\
CHEMDNER patent & 0.6828 & 0.8382 & 0.7526 && 0.6875 & 0.8423 & 0.7570 && 0.0047 & 0.0041 & 0.0044\\
IEPA & 0.8771 & 0.8771 & 0.8771 && 0.8754 & 0.8870 & 0.8812 && -0.0017 & 0.0099 & 0.0041\\
JNLPBA & 0.8366 & 0.8561 & 0.8462 && 0.8287 & 0.8507 & 0.8396 && -0.0079 & -0.0054 & -0.0066\\
LocText & 0.8646 & 0.8202 & 0.8418 && 0.8689 & 0.8881 & 0.8784 && 0.0043 & 0.0679 & 0.0366\\
miRNA & 0.7644 & 0.7956 & 0.7797 && 0.7541 & 0.8679 & 0.8070 && -0.0103 & 0.0723 & 0.0273\\
OSIRIS & 0.8721 & 0.8926 & 0.8823 && 0.9123 & 0.9430 & 0.9274 && 0.0402 & 0.0504 & 0.0451\\
Variome & 0.9223 & 0.9482 & 0.9351 && 0.9169 & 0.9519 & 0.9340 && -0.0054 & 0.0037 & -0.0011\\
\midrule
 avg. & 0.8267 & 0.8245 & 0.8242 && 0.8376 & 0.8586 & 0.8470 && 0.0150 & 0.0362 & 0.0240\\
 & & & && & & \\
Species & & & && & & \\
\cmidrule{1-1}
CellFinder & 0.8489 & 0.9219 & 0.8839 && 0.8414 & 0.9531 & 0.8938 && -0.0075 & 0.0312 & 0.0099\\
CHEBI & 0.8875 & 0.7890 & 0.8353 && 0.8807 & 0.7765 & 0.8253 && -0.0068 & -0.0125 & -0.0100\\
Linneaus & 0.9440 & 0.9142 & 0.9289 && 0.9579 & 0.9470 & 0.9524 && 0.0139 & 0.0328 & 0.0235\\
LocText & 0.9545 & 0.9130 & 0.9333 && 0.9468 & 0.9674 & 0.9570 && -0.0077 & 0.0544 & 0.0237\\
miRNA & 0.9914 & 0.9312 & 0.9603 && 0.9789 & 0.9393 & 0.9587 && -0.0125 & 0.0081 & -0.0016\\
S800 & 0.7664 & 0.7232 & 0.7442 && 0.7396 & 0.7518 & 0.7457 && -0.0268 & 0.0286 & 0.0015\\
Variome & 0.5400 & 0.8182 & 0.6506 && 0.6829 & 0.8485 & 0.7568 && 0.1429 & 0.0303 & 0.1062\\
\midrule
 avg. & 0.8475 & 0.8587 & 0.8481 && 0.8612 & 0.8834 & 0.8700 && 0.0312 & 0.0283 & 0.0252\\
\bottomrule
\end{tabular}

\end{table*} 

\clearpage

\bibliographystyle{natbib}
\bibliography{literature}